# Federated Isolation Forest for Efficient Anomaly Detection on Edge IoT Systems

Pavle Vasiljevic
*University of Novi Sad*
*Faculty of Technical Sciences*
Novi Sad, Serbia
pavle.vasiljevic@uns.ac.rs

Milica Matic
*University of Novi Sad*
*Faculty of Technical Sciences*
Novi Sad, Serbia
milica.matic@rt-rk.com

Miroslav Popovic
*University of Novi Sad*
*Faculty of Technical Sciences*
Novi Sad, Serbia
miroslav.popovic@rt-rk.uns.ac.rs

***Abstract*— Recently, federated learning frameworks such as Python TestBed for Federated Learning Algorithms and MicroPython TestBed for Federated Learning Algorithms have emerged to tackle user privacy concerns and efficiency in embedded systems. Even more recently, an efficient federated anomaly detection algorithm, FLiForest, based on Isolation Forests has been developed, offering a low-resource, unsupervised method well-suited for edge deployment and continuous learning. In this paper, we present an application of Isolation Forest-based temperature anomaly detection, developed using the previously mentioned federated learning frameworks, aimed at small edge devices and IoT systems running MicroPython. The system has been experimentally evaluated, achieving over 96% accuracy in distinguishing normal from abnormal readings and above 78% precision in detecting anomalies across all tested configurations, while maintaining a memory usage below 160 KB during model training. These results highlight its suitability for resource-constrained environments and edge systems, while upholding federated learning principles of data privacy and collaborative learning.**

## I. INTRODUCTION

This research is conducted within the ongoing EU Horizon 2020 project entitled Trustworthy and Resilient Decentralized Intelligence for edge Systems (TaRDIS) [1]. Main objective of this project is to develop a toolbox for intuitive programing of decentralized and distributed applications, primarily in the edge systems, including but not limited to privacy preserving federated learning in smart homes, highly resilient industrial internet of things (IoT) applications and electrical vehicle smart grids.

The task of temperature anomaly detection in small IoTs can be quite challenging due to the limited amount of resources available and possibly fast changing temperature conditions, that can vary based on temperature position, exposure to heating bodies and other outside factors. This task finds many uses, in home safety systems with early fire detection, healthcare with patient monitoring and industrial uses, such as machine health monitoring and cold storage monitoring, preventing spoilage of goods.

Federated learning [2] enhances privacy by allowing users to benefit from model improvements without sharing their sensor data, which is crucial for those concerned about data security. This method is especially useful for problems with highly variable data, such as temperature anomaly detection, where readings differ significantly based on the environment of the IoT devices. Additionally, it improves efficiency by reducing the need for constant data transmission, lowering bandwidth usage, and enabling local processing on edge devices.

Python TestBed for Federated Learning Algorithms (PTB-FLA) [3] and its successor, MicroPython TestBed for Federated Learning Algorithms (MPT-FLA) [4], aimed at MicroPython enabled embedded devices deployment, have been used for development of centralized, decentralized and peer-to-peer data exchange used in time division multiplexing communication based federated learning (FL) applications. They leverage the Single Program Multiple Data (SPMD) pattern, to provide simple development experience to ML and AI developers that are not as experienced with distributed systems development.

Another strong suit of the two frameworks is that the centralized and decentralized generic algorithms that are key to the way they are used have been formally verified using CSP [5], ensuring deadlock freeness and successful termination. In addition to this the development process from sequential machine learning code to federated code has been standardized with the use of PTB-FLA federated learning development paradigm [6].

Isolation forests [7] offer a good balance between efficiency and performance, while also having a benefit of being an unsupervised anomaly detection method which is crucial for edge deployment. They detect anomalies by recursively partitioning data points using random splits, isolating outliers in fewer steps than normal instances due to their distinctiveness. The downside of this approach is that model aggregation is nontrivial and requires careful implementation.

Lately a lot of research has been done looking to improve the efficiency of isolation forests in anomaly detection. One approach focuses on improving the original Isolation Forest algorithm by modifying the tree construction to retain dense regions of normal data which allows better clustering of

Funded by the European Union (TaRDIS, 101093006). Views and opinions expressed are however those of the author(s) only and do not necessarily reflect those of the European Union. Neither the European Union nor the granting authority can be held responsible for them.

normal data while still effectively identifying outliers which provides a more interpretable model [8]. Another approach introduces a multi-level subspace partitioning technique, refining anomaly scores by considering both global and local perspectives, thereby improving detection accuracy for complex data distributions [9]. Additionally, recent work has explored the integration of Federated Learning with Isolation Forest to enable privacy-preserving anomaly detection in IoT networks, allowing decentralized data processing without compromising detection performance [10].

Quite recently a new algorithm called FLiForest [11] has emerged adapting the Isolation Forest Algorithm for use in federated learning, solving the isolation forest aggregation by training the model in a layer-by-layer manner while sharing only split values and using the global model for client inference.

The main original contributions of this paper are: (1) the specialization of the algorithm FLiForest based on PTB-FLA (called PFLiForest), (2) the PFLiForest feasibility assessment for temperature anomaly detection in IoT devices, (3) the PFLiForest experimental evaluation. The PFLiForest has been experimentally evaluated using temperature data obtained from real world sensors.

The rest of the paper is organized as follows. Section 2 presents the system design, Section 3 comprises the systems experimental evaluation, and Section 4 concludes the paper.

## II. SYSTEM DESIGN

### A. System Architecture

System follows a centralized federated learning scenario comprising of an edge server and *n* clients running on small IoTs with temperature sensors on them (e.g., RPi Pico W).

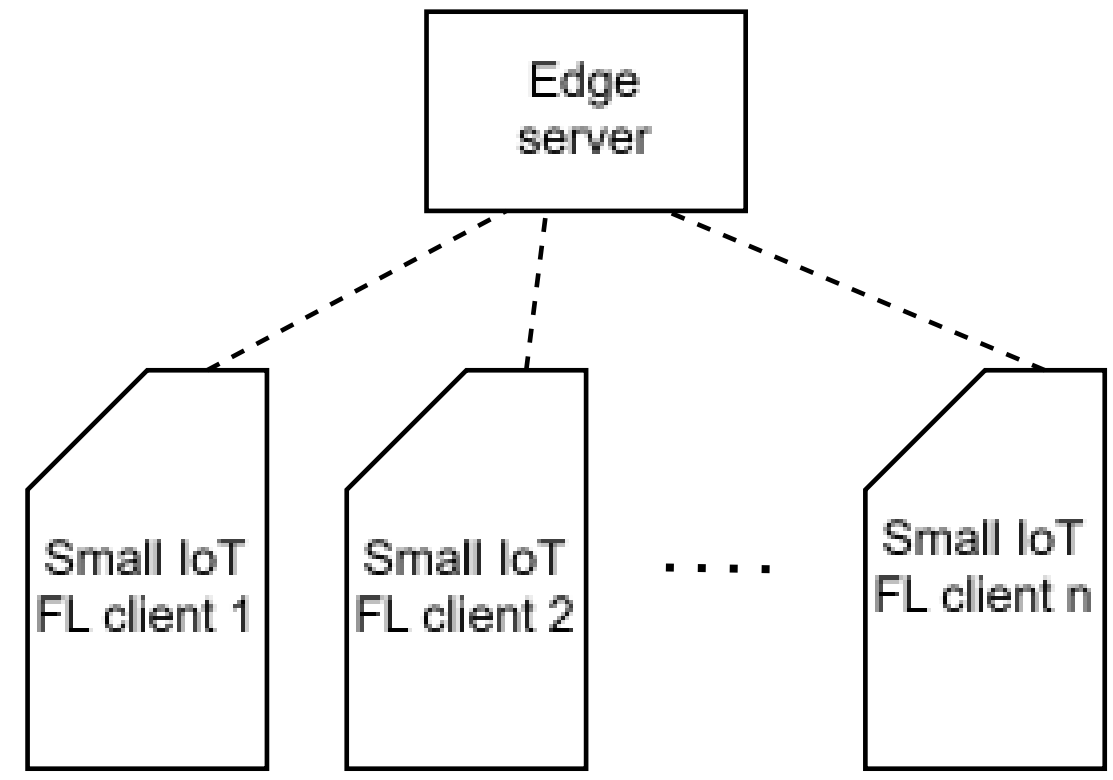

Fig. 1. Federated anomaly detection system architecture

The nodes communicate via PTB-FLA testbed instances running on each of them. The function fl_centralized implements a generic centralized federated learning algorithm using callback functions, following the Single Program Multiple Data (SPMD) pattern.

### B. System Behaviour

Due to the layer-by-layer nature of the way the model training is conducted in the algorithm FLiForest, the traditional approach of training the whole models inside the client callback functions was divided such that: (i) the client callback functions calculate values at which the tree is supposed to be split and (ii) the edge server callback function aggregates them.

The PFLiForest tree-building algorithm (Algorithm 1) is implemented iteratively, using a queue (deque) to manage tree growth instead of relying on the call stack. This modification maintain compatibility with the PTB-FLA framework. Additionally, an FL phase synchronization mechanism was introduced to ensure all nodes remain aligned during training. Without synchronization, some nodes might complete the construction of a single tree earlier than others and prematurely start building the next tree in the forest. Such asynchronism would contradict the PTB-FLA function fl_centralized, where a new iteration of model training does not begin until the global model is disseminated to all client nodes by the server.

After initializing tree construction variables, the function iteratively pops elements from the left side of the queue, performs centralized federated learning, and ensures phase synchronization. If clients have not yet finished building their tree, new nodes are created, data is split into new partitions, and these partitions are enqueued for further processing until the tree is fully constructed.

```
Globally defined variables used in all the algorithms
01: //current phase
02: SERVER_PHASE = 0
03: CLIENT_PHASE = 0
04: //states
05: INITIAL = 0
06: CLIENT_RESTING = 1
07: END_TREE = 2
```

```
Algorithm 1. Build isolation tree
01: build_iTree (ptbFla, data, max_depth):
02:   // Initialize tree construction variables
03:   global CLIENT_PHASE, SERVER_PHASE
04:   root = IsolationTreeNode()
05:   queue = deque([(root, data, 0)])
06:   while True:
07:     if queue:
08:       //get values from the queue
09:       node, pData, depth = queue.popleft ()
10:     else:
11:       node, pData, depth = None, None, None
12:     //Perform federated learning communication
13:     ret = ptbFla.fl_centralized (serverProcessing,
              clientProcessing, localData, pData, 1)
14:     //Perform phase sync
15:     phase = ptbFla.fl_centralized (serverPhase,
              clientPhase, ret["phase"], ret["phase"], 1)
16:     //if all the clients are done building their tree exit
17:     if phase == END_TREE:
18:       //return the completed tree
19:       return root
20:     //continue to the next iteration if server node
21:     if ptbFla.nodeId = ptbFla.flSrvId:
22:       continue
23:     // if the client is in the resting phase, stop, don't add
          more nodes
24:     if CLIENT_PHASE == CLIENT_RESTING or
        not node:
25:       continue
26:     // put client into a resting phase when it's done
          building a tree
```

```
27:    if depth >= max_depth or len(set(pData)) <= 1:
28:      if not queue:
29:        CLIENT_PHASE = CLIENT_RESTING
30:      continue
31:    //make new left and right nodes
32:    node.left = IsolationTreeNode()
33:    node.right = IsolationTreeNode()
34:    //assign the split value
35:    node.split_value = ret["data"]
36:    //split data into left and right partitions
37:    left_partition, right_partition =
                              split_data(node.split_value)
38:    //append nodes, partitions and
39:     //incremented current depth to queue
40:    queue.append((node.left, left_partitions,
                                         depth + 1))
41:    queue.append((node.right, right_partitions,
                                         depth + 1))
```

Algorithm 2 builds an isolation forest in a for loop (line 3) by calling Algorithm 1 (line 4) to get the next *tree* and then add it to the *forest* (line 5). The for loop is repeated until the desired number of trees (*num_trees*) is reached. At the end, Algorithm 2 returns the created isolation forest model (line 6).

```
Algorithm 2. Build isolation forest
01: build_iForest (ptbFla, data, num_trees, max_depth):
02:   forest = []
03:   for _ in range(num_trees)
04:     tree = build_iTree (ptbFla, data, max_depth)
05:     forest. append(tree)
06:   return forest
```

Algorithm 3 represents the client callback function, which serves two key roles. Its primary role is to generate client-specific split points using private data. The secondary role is to notify the server when the client has completed building its tree by signalling its transition to the resting phase.

```
Algorithm 3. Client processing
01: clientProcessing (localData, privateData, msg):
02:   global CLIENT_PHASE
02:   // relaying that client is done building its tree
03:   if CLIENT_PHASE == CLIENT_RESTING:
04:     return {"phase": CLIENT_PHASE, "data": 0}
05:   // Compute client-side splits
06:   splits = client_process_layer(privateData)
07:   // return the client splits
08:   return {"phase": CLIENT_PHASE, "data": splits}
```

Algorithm 4 represents the server callback function, where client splits are aggregated by averaging their values. These global values are then distributed to individual clients to form the current layer, eventually constructing a complete isolation tree. Additionally, the server callback determines when to conclude the tree-building process by filtering out messages from clients in the resting phase. Once all clients enter this phase, all messages are filtered out, and the server signals the completion of tree building to all clients.

```
Algorithm 4. Server processing
01: serverProcessing (localData, msgs):
02:   global SERVER_PHASE
03:   clientSplits = []
03:   // Filtering messages of clients who are resting
04:   for item in msgs:
05:     if item["phase"] != CLIENT_RESTING:
06:       clientSplits.append(item["data"])
07:   //if there are no client splits for aggregation, then all
          clients are done constructing their trees, can move
          to the next tree
08:   if length(clientSplits) == 0:
09:     SERVER_PHASE = END_TREE
10:     return {"phase": SERVER_PHASE, "data": None}
11:   // Aggregate client data
12:   return {"phase": SERVER_PHASE, "data":
              server_aggregate_layer (clientSplits)}
```

The phase synchronization procedure in line 12 of Algorithm 4 is crucial to the tree construction process, ensuring the proper dissemination of the signal END_TREE, which serves as the exit condition for the tree-building algorithm. The callback function serverPhase transmits the server's phase (either INITIAL_SERVER or END_TREE) to the clients via PTB-FLA's generic algorithm fl_centralized. The callback function clientPhase then receives this phase and uses it to determine the exit condition for the client's tree building process.

The trained isolation forest model is represented as a list of isolation trees, where each tree is constructed of objects of the class IsolationTreeNode, see Fig. 2.

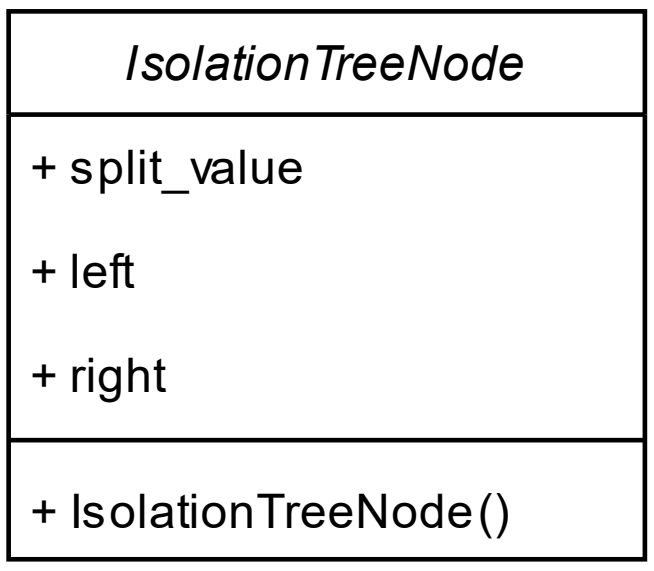

Fig. 2. The UML class diagram for the class IsolationTreeNode

### C. System inference

The isolation forest inference is done by computing an anomaly score for a given data point by traversing the isolation trees and averaging path lengths across an isolation forest. The anomaly score is computed as:

$$s(x,n) = 2^{-\frac{E(h(x))}{c(n)}} \tag{1}$$

$$c(n) = 2(ln(n-1) + 0.577) - \frac{2(n-1)}{n} \tag{2}$$

Where *E(h(x))* is the average path length across the isolation forest, and *c(n)* in (2) is the normalization factor dependent on the size of the dataset the tree was trained on. The calculated anomaly score is then compared against an anomaly threshold to classify a value as either an anomaly or normal.

## III. Experimental Evaluation

In this section we present the PFLiForest experimental evaluation results and their discussion.

Dataset used for performance measurement is the set of the temperature values in Celsius. The model's performance is evaluated using the metrics: Area Under the Receiver Operating Characteristics Curve (AUC-ROC) and Area Under the Precision-Recall Curve (AUC-PR).

The primary goal of this research is to determine the optimal isolation forest that fits the memory footprint of a target embedded device The target device is the Raspberry PI Pico microcontroller featuring a dual-core ARM Cortex-M0 processor with 256 KB internal RAM and support for up to 16 MB of off-chip flash storage.

To conclude this section, we provided a performance comparison between our PFLiForest and the original algorithm called Isolation Forest (iForest) [7] that is used here as a baseline.

### A. *Experimental Setup*

The evaluation of the anomaly detection algorithm PFLiForest is based on experiments with the following parameter values: the isolation forest size {10, 25, 50, 75 ,100} trees; the maximum tree depth {4, 6, 8, 10 ,12}; the training dataset size {50, 100, 150, 200}. To minimize variability, each experiment was conducted 40 times. Iteration to iteration data in the training datasets remained unchanged. After each iteration, the training dataset was used to generate a new synthetic testing dataset by randomly sampling from the original data with added Gaussian noise. Additionally, anomalies were introduced in 10% of the data by replacing selected values with random values beyond the original data range. During the test data generation process, we also manually added the corresponding labels, used for subsequent performance evaluation. After each iteration the model parameters and performance metrics were stored in a database, which was later queried to produce averages across the corresponding model parameters.

### B. *Evaluation metrics*

In our experiments we are using the AUC-ROC and the AUC-PR as the evaluation accuracy criteria. Both AUC-ROC and AUC-PR are widely used to ensure that the model performs well. They both provide a graphical representation of the results.

A confusion matrix is a 2x2 matrix, presented in the TABLE I, shows how well the classification model performs combining the predicted and actual positive and negative values:

- True Positive (TP) – the number of correctly predicted positive samples.
- False Positive (FP) – the number of samples incorrectly predicted as positive.
- False Negative (FN) – the number of samples incorrectly predicted as positive.
- True Negative (TN) - the number of correctly predicted negative samples.

Total number of samples *S* can be calculated as:

$$S = TP + FP + FN + TN \tag{3}$$

TABLE I. CONFUSION MATRIX

| | | Actual | |
|---|---|---|---|
| | | Positive | Negative |
| Predicted | Positive | **True Positive** | **False Positive** |
| | Negative | **False Negative** | **True Negative** |

From these values, the following indicators can be calculated:

- True Positive Rate (TPR) – the ratio of the correctly predicted positive samples; this metric is also known as recall or sensitivity
- False Positive Rate (FPR) – the ratio of the incorrectly predicted positive samples.
- Positive Predictive Value (PPV) – the ratio of positive samples correctly predicted as positive; this metric is also known as precision.

$$TPR = \frac{TP}{TP+FN} = \frac{correctly\ predicted\ positive\ samples}{all\ samples\ classified\ as\ positive} \tag{4}$$

$$FPR = \frac{FP}{FP+TN} = \frac{negative\ samples\ predicted\ as\ positive}{all\ actual\ negative\ samples} \tag{5}$$

$$PPV = \frac{TP}{TP+FP} = \frac{correctly\ predicted\ positive\ samples}{all\ samples\ clasified\ as\ positive} \tag{6}$$

$$F1_{score} = 2\frac{TPR \times PPV}{TPR+PPV} = 2\frac{recall \times precision}{recall+precision} \tag{7}$$

Precision value shows how well the model forecasts the positive outcomes. A high precision value indicates the model has a low rate of false positives.

The recall shows how well the model can capture each relevant instance. A high recall value indicates that the model has a low rate of false negatives.

The ROC curve plots the TPR against the FPR. AUC-ROC represents the area under the ROC curve. The PR curve plots the PPV (Precision) against the TPR (Recall). The AUC-PR represents the area under the PR curve.

To determine the optimal threshold for anomaly classification, we will utilize the F1-score since it balances precision and recall leading to the best overall model performance.

Along with these model performance metrics, during the experimental validation we also recorded the maximum memory usage while training and training time. These metrics present a way to gauge model efficiency and determine whether its suitable for use in devices with limited system resources.

### C. *Results and Discussion*

The evaluation parameters considered in evaluations are:

- The maximum depth of the tree (4, 6, 8 and 10)
- The number of trees in the forest (10, 25, 50, 75)
- The amount of the training data (50, 100, 150, 200).

To present the evaluation results, 4 points were selected to represent a broad range of parameter combinations. Each point represents a triple of (*max_depth*, *number_of_trees*, *training_data_amount*).

The 4 chosen points are:

- A (4, 10, 50)
- B (6, 25, 100)

- C (8, 50, 150)
- D (10, 75, 200).

*1) Memory usage against AUC-ROC and AUC-PR*

Fig. 3. shows AUC-ROC and AUC-PR against the memory usage values for the 4 points A, B, C, and D.

The goal is to fit the model into the RPi Pico device. Therefore, we search for point E with the maximum values of AUC-ROC and AUC-PR where memory usage is ~160 kB:

$$E = M(\max(AUC_{ROC}), \max(AUC_{PR})) < 160 \quad (8)$$

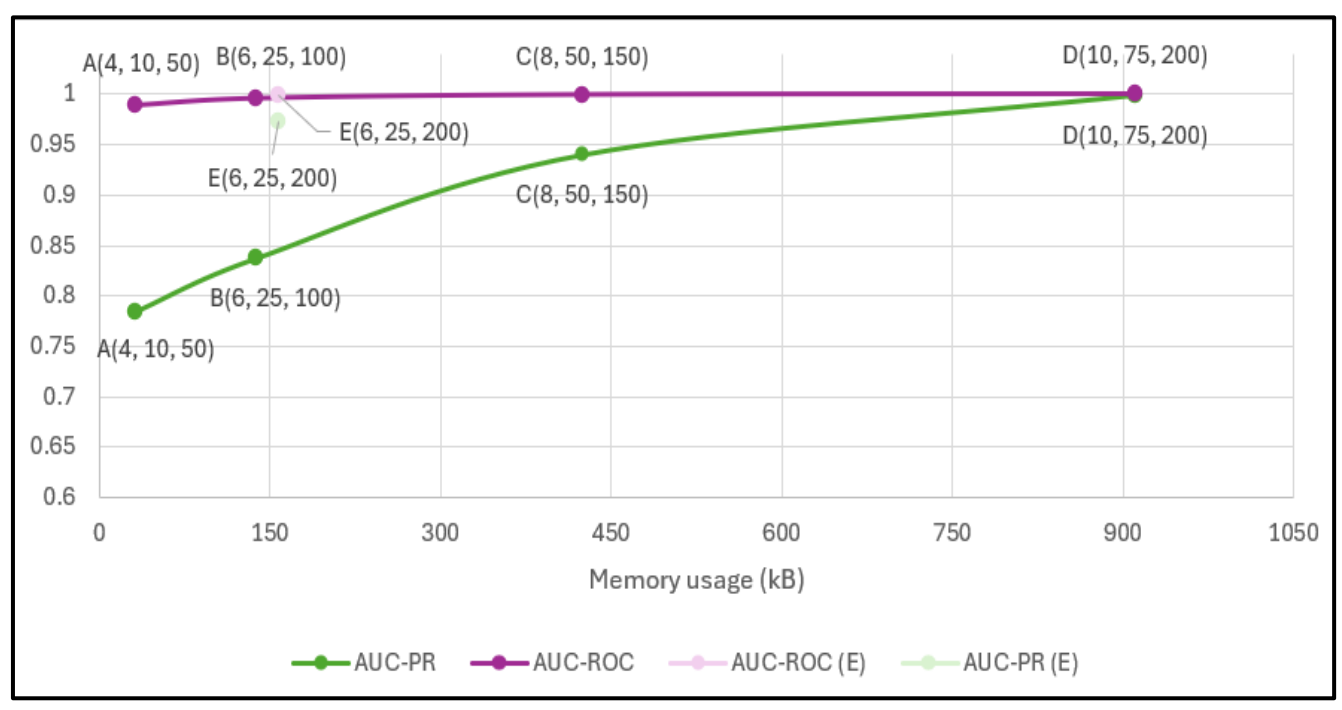

Fig. 3. AUC-ROC and AUC-PR versus the memory usage (the model size)

The chosen point E (6, 25, 200) has AUC-ROC value of 0.9996 and AUC-PR of 0.9728 while using 153 kB of memory for training, which is confidently within the desired memory usage.

*2) Execution Time Against Memory Usage*

The execution time is the time taken to build a forest in seconds. Fig. 4. shows the execution time as a function of the memory usage i.e., the model size in the given points. As expected, the execution time is almost a linear function of the model size (quantified as the memory usage).

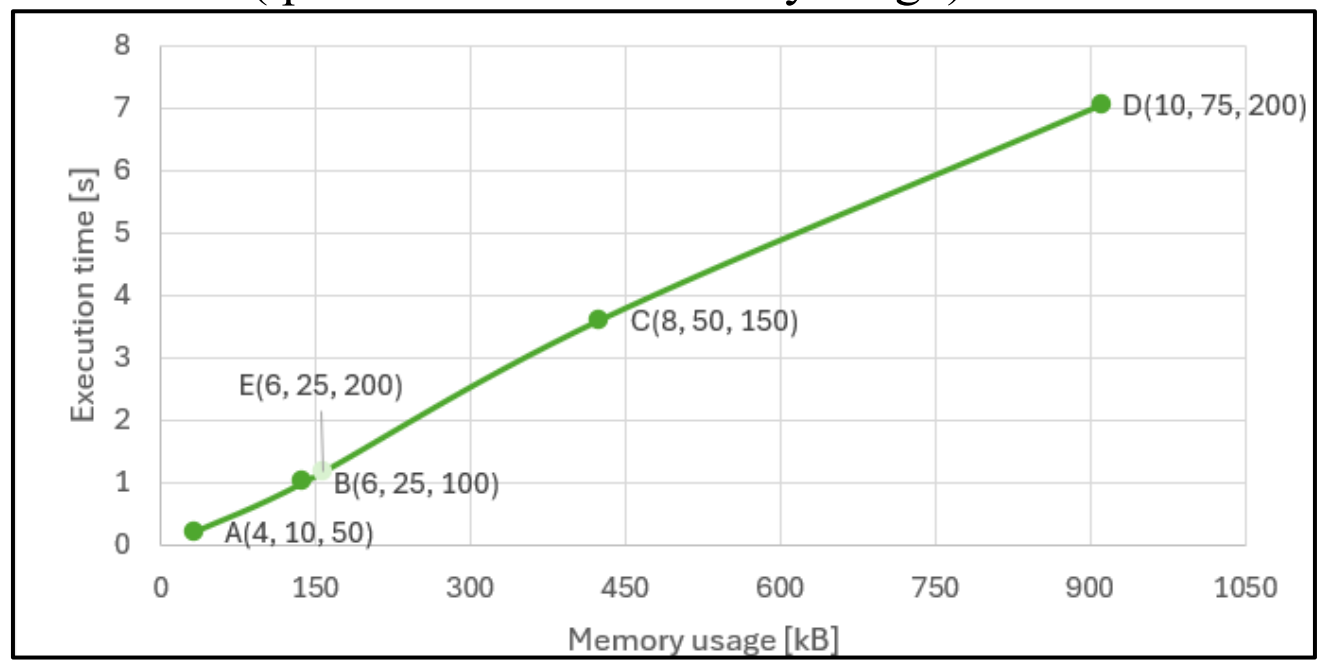

Fig. 4. The execution time versus the memory usage i.e., the model size

Going back to Fig 3., it can be concluded that the growth of the tree depth, number of the trees in the forest and amount of the training data directly cause the increase in memory usage. Point E, chosen in test 1, passes the execution time test as well, since an increase in the amount of training data did not affect memory usage.

*3) Further performance analysis for the chosen point*

The chosen point E (6, 25, 200) demonstrated a good balance between model performance in terms of AUC-ROC and AUC-PR, as well as resource efficiency. To further validate this choice, we evaluated it again by constructing a confusion matrix in TABLE II and calculating the true positive rate, false positive rate, and the positive predictive value.

TABLE II. CONFUSION MATRIX FOR THE CHOSEN POINT E (6, 25, 200)

| | | Actual | |
|---|---|---|---|
| | | Positive | Negative |
| Predicted | Positive | **998.0** | **8.325** |
| | Negative | **2.0** | **8991.675** |

The experiment was again conducted 40 times on a synthetic dataset containing 10000 values with 10% of the data being anomalies. The anomaly threshold value for evaluation (0.8265) was chosen by picking a value on the precision recall curve where an F1-Score was the highest.

In the additional evaluation for the chosen point, with the optimal anomaly threshold value, the model had a very high precision (99.17%) and recall (99.8%), along with a FPR below 0.1%, which was to be expected for high AUC-ROC and AUC-PR values.

*4) Comparison to the base case*

Here we give a comparison between our PFLiForest and the original iForest. Fig. 5. presents the AUC-PR as the function of the model size (i.e., the memory usage). AUC-ROC values are omitted from the graph, as they are nearly identical for both algorithms, consistently exceeding 96% across all parameter configurations and reaching 99% for the vast majority of cases.

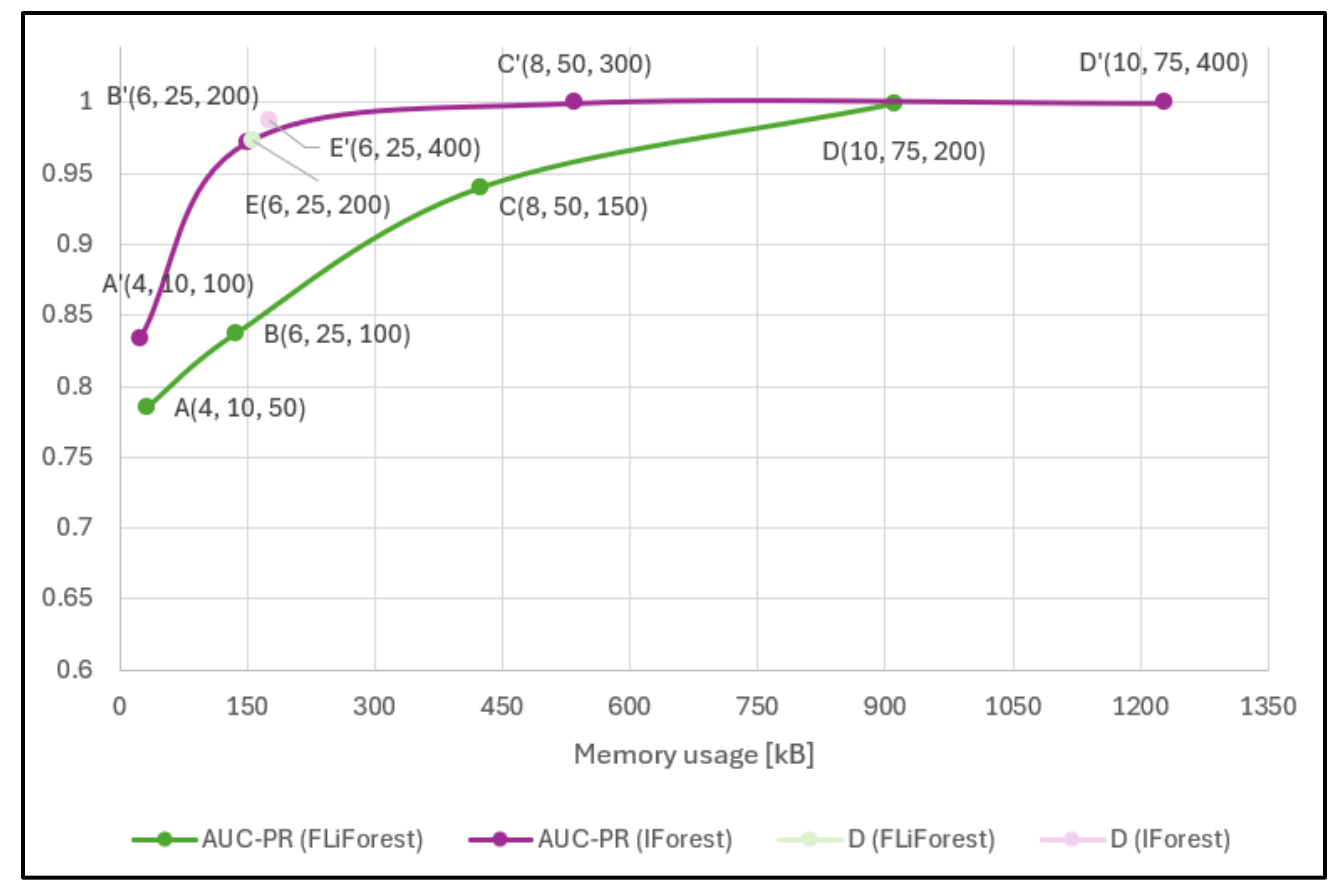

Fig. 5. AUC-PR of PFLiForest and iForest versus the memory usage

Calculations are given for the points A, B, C, and D for PFLiForest and the corresponding points A' (4, 10, 100), B' (6, 25, 200), C' (8, 50, 300), and D' (10, 75, 400) for iForest. The point E and the corresponding point E' are once again proven to satisfy the condition (8). Moreover, the points E and E' are close in both memory usage and model performance further solidifying the point E as the prime candidate for deployment to the target device. The lower overall AUC-PR of PFLiForest compared to iForest stems from its federated nature, reflecting a trade-off that enhances data privacy by preventing direct data sharing. However, this gap narrows as the memory usage i.e., the model size grows, eventually converging with the baseline performance.

## IV. CONCLUSION

In this paper, we present the PFLiForest, a federated temperature anomaly detection algorithm targeting embedded devices, such as small IoTs. The PFLiForest is the algorithm FLiForest specialization based on the PTB-FLA framework. We also carried out a comprehensive experimental evaluation to assess both the performance and efficiency of the system, along with a feasibility analysis of deploying PFLiForest for temperature anomaly detection in resource-constrained environments.

The main advantages, of this approach are the following: (i) it achieves relatively high system performance while keeping memory and CPU usage low, provided the parameters are appropriately tuned, (ii) it enables context sharing between clients, leading to better generalization, (iii) it allows for additional training and adaptation to current environmental conditions, enabling anomaly detection without requiring data labeling or human interaction—making it ideal for smart IoT devices.

The main shortcomings of the current system are as follows: (i) high communication cost and potential for bursty traffic during the training of the federated isolation forest, (ii) lower overall model performance compared to the sequential algorithm iForest (using the same parameters), primarily due to the smaller amount of data available to each client during training.

In our future work we plan on doing a comparative study of the PFLiForest evaluated in this paper and the algorithm proposed by Li [10]. Another interesting research direction could be exploring the vertical federated learning for isolation forests, focusing on secure split aggregation, efficient tree construction across feature-partitioned clients, and its impact on anomaly detection performance in distributed environments.